\documentclass[letterpaper, 10 pt, conference]{ieeeconf}  

\IEEEoverridecommandlockouts                              

\title{\LARGE \bf
A Unified Framework for Multi-Contact Path Planning in the Rolling Robot Systems
}

\author{Qing Yu$^{1}$, Mikhail Svinin$^{2}$ and Seyed Amir Tafrishi$^{1}$
\thanks{$^{1}$Qing Yu and Seyed Amir Tafrishi are with Geometric Mechanics and Mechatronics in Robotics (gm$^2$R) Lab, School of Engineering, Cardiff University, Queen's Buildings, The Parade, Cardiff, CF24 3AA 
\tt\small \{YuQ16, tafrishisa\}@cardiff.ac.uk}
\thanks{$^{2}$Mikhail Svinin is with Information Science and Engineering Department, Ritsumeikan University, Shiga, Japan   
{\tt\small svinin@fc.ritsumei.ac.jp}}
\thanks{*Supported by China Scholarship Council (No.202006760092)}
\thanks{Seyed Amir Tafrishi is the corresponding author of this study (phone: +44 29208 76176, e-mail: {\tt\small Tafrishisa@cardiff.ac.uk}).}
}

\usepackage{subcaption}
\usepackage{mathtools}
\usepackage{nomencl}

\usepackage{lipsum}     
\usepackage{graphicx}
\usepackage{algpseudocode}

\usepackage{algorithm}
\usepackage{changes}
\usepackage{xpatch}
\usepackage{multicol}
\usepackage{graphics}
\usepackage{cite}
 \usepackage{amssymb}
 
\usepackage{bm}
\usepackage{amsmath,amsfonts}

\DeclareRobustCommand{\uvec}[1]{{%
		\ifcsname uvec#1\endcsname
		\csname uvec#1\endcsname
		\else
		\bm{\mathbf{#1}}%
		\fi
}}
\usepackage{algpseudocode}
\usepackage{algorithm}
\usepackage{algpascal}
\usepackage[english]{babel}
\usepackage{float}
\usepackage[colorlinks]{hyperref}

\hypersetup{
    colorlinks=true,
    linkcolor=orange,
    filecolor=green, 
    urlcolor=black,
    citecolor=violet
}

\renewcommand{\baselinestretch}{.92}
\begin{document}

\maketitle
\thispagestyle{empty}
\pagestyle{empty}
\begin{abstract} 
Rolling motion planning is challenging because rolling contact imposes nonholonomic constraints and the configuration evolves on a curved manifold. The problem becomes substantially harder in multi-contact settings, where multiple bodies roll without slip and the contact states are coupled. This paper presents a new framework for multi-contact path planning in spherical rolling robotics under no-slip constraints. We first derive a compact kinematic model for multi-sphere rolling using Montana’s contact-coordinate formulation, where each contact is represented by a stacked five-state vector. Building on this model, we construct a Voronoi-based roadmap directly on the spherical contact manifold, incorporating spherical-cap obstacles and mutual-exclusion regions via on-manifold collision checking, and refine discrete graph paths using manifold-consistent log–exp smoothing. The resulting smoothed surface paths are then lifted to admissible multi-contact rolling motions through the derived Montana kinematics and validated via forward simulation. We further evaluate feasibility and path quality versus trajectory smoothness, Voronoi seed density, and computation time. The proposed framework provides a foundation for extending the method to non-spherical geometries, time-varying obstacle environments, and experimental validation on physical rolling robotic platforms.
\end{abstract}

\section{Introduction} 
Rolling contact is a key mechanism for generating constrained, repeatable motion in robotics and underpins a growing range of applications \cite{Montana1988,NonprehensilePlanningPlane2019,tafrishi2025survey}, notably spherical reconfigurable robots \cite{zhong2022kin,Wiltshire2024disknovel,reconfigurable} and dexterous manipulation/grasping \cite{tahara2012externally,yuan2020designICRA,dexterousmanu}. Under a no-slip assumption, rolling yields nonholonomic velocity constraints that couple rotation and translation through contact geometry \cite{Montana1988}, offering a more structured interaction than sliding and enabling precise control and coordinated behaviours \cite{Li1990,tafrishi2025survey}. In reconfigurable rolling robots, this supports module traversal over a host surface while maintaining continuous engagement \cite{509415,Svinin2008}; in manipulation, fingertip rolling enables stable reorientation without releasing the grasp \cite{Murray1994,tahara2012externally,yuan2020designICRA}. However, with multiple simultaneous rolling contacts, planning becomes markedly harder: coupled constraints must be satisfied while performing obstacle avoidance in 3D environments \cite{tafrishi2025survey}, and a clear, widely adopted framework for multi-contact rolling path planning is still missing.

\begin{figure}[t!]
	\centering
	\includegraphics[width=3.5 in]{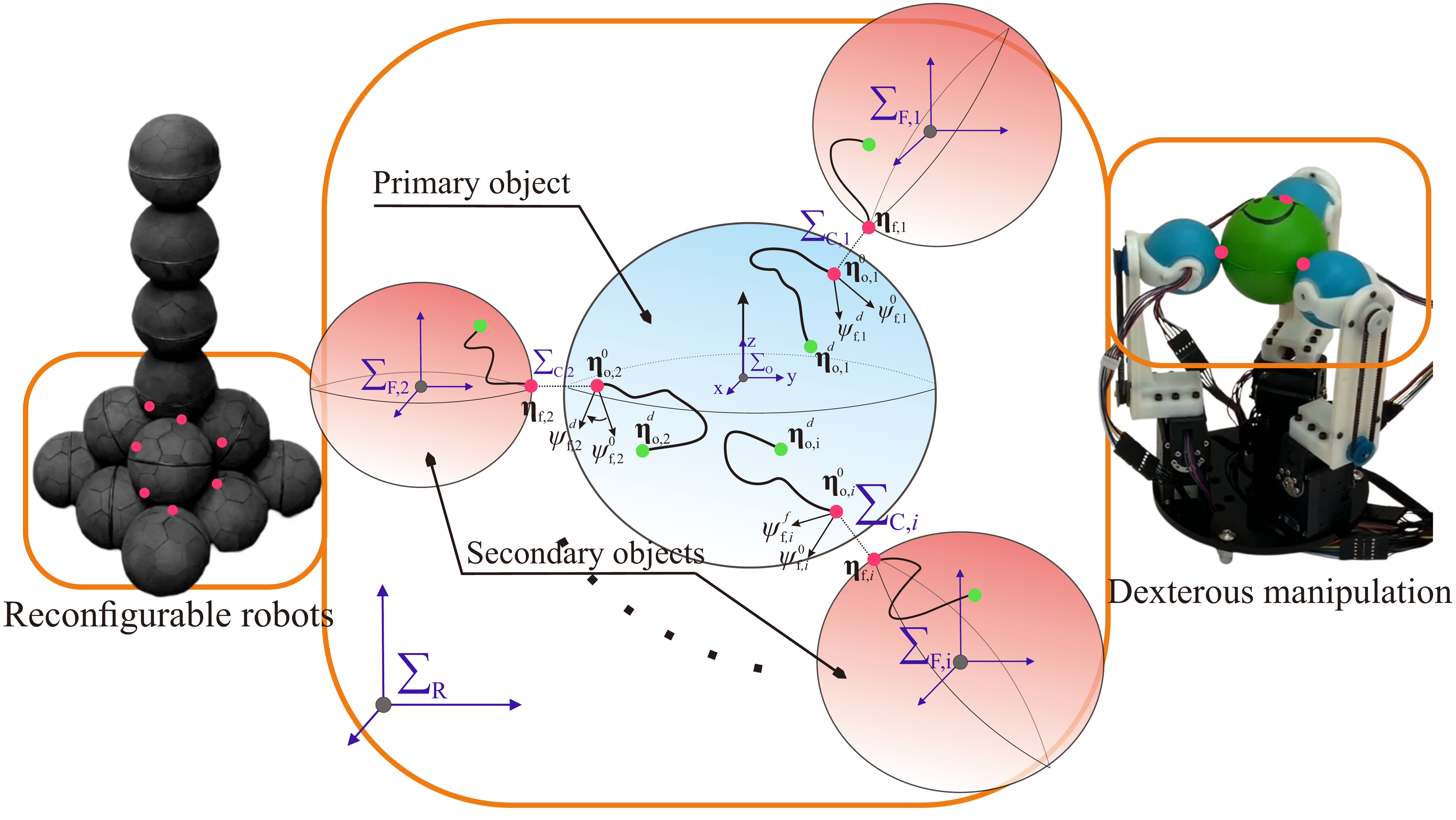}
	\caption{Multi-contact spherical rolling model. A primary sphere $U_o$ (blue) is contacted by $i$ secondary rolling spheres $U_{f,i}$ (red), motivated by reconfigurable rolling robots and rolling-contact dexterous manipulation (examples left/right). Each contact $i$ is parameterized by a local frame $\Sigma_{C,i}$, surface coordinates $(u_{o,i},v_{o,i})$ and $(u_{f,i},v_{f,i})$, and the relative spin angle $\psi_i$, all referenced to the inertial frame $\Sigma_R$.}
 \label{Fig:IROSQing}
\end{figure}

Multiple rolling contacts often coexist, extending the problem from single-contact interactions to coupled multi-body systems. Historically, rolling-contact kinematics progressed from curvature-based formulations (e.g., Woronetz’s early analysis \cite{MathematischeAnnalen1911}) to the widely adopted Montana contact equations, which express local contact-state rates directly through geometric operators and relative body motion \cite{MathematischeAnnalen1911,Montana1988}. While single-contact rolling is now well studied—particularly in the canonical ball–plate setting \cite{svinin2008motion,tafrishi2023geometric,tafrishi2025survey}—multiple simultaneous contacts introduce strong coupling: the nonholonomic velocity constraints become interdependent across contacts and bodies, reducing maneuverability and making admissible motion computation and coordination substantially more difficult \cite{Chitour2012,Cui2010}. Early multi-contact studies examined trapped-object manipulation with multiple contacts (e.g., two-contact formulations and multi-finger abstractions) \cite{sarkar1997control,kiss2002motion}, whereas more recent work has revisited multi-body rolling by systematically assembling multi-contact models from Montana-type building blocks and clarifying how each local contact state depends on the global motion of the overall system \cite{tafrishi2025survey}. This challenge is central to spherical reconfigurable robots \cite{Wiltshire2024disknovel,zhong2022kin}, where multiple modules roll around a host sphere, and to rolling-contact grasping mechanisms \cite{tahara2012externally,yuan2020designICRA} that reorient a trapped object via free-rolling fingertips (see Fig.~\ref{Fig:IROSQing}).

Motivation and exciting problems  (multi-contact path planning - obstacle avoidance)However, when multiple rolling contacts coexist, the associated kinematic equations become strongly coupled across bodies. Directly incorporating these coupled multi-contact kinematics into path planning, therefore, introduces significant computational complexity, particularly when obstacle avoidance must be resolved simultaneously \cite{Sarkar1994,Papadopoulos1994}. This difficulty is consistent with broader observations in modern path-planning surveys for ground robots and UAVs, where collision avoidance and coordination in cluttered, often dynamic environments remain major scaling bottlenecks \cite{Liu2023ESWA,Jones2023UAVSurvey}. In multi-body rolling on a host surface, the challenge is even more restrictive: admissible motions must remain on a curved $2$D manifold embedded in $\mathbb{R}^3$, and the usable contact workspace can be partially occluded by other rolling bodies, reducing feasible clearance compared to free-flying agents. These challenges motivate a planning strategy that explicitly captures geometric separation before enforcing rolling-consistent kinematics. Voronoi-based geometric reasoning offers a structured representation of spatial separation on constrained surfaces, naturally promoting clearance from obstacles \cite{Choset2005,Aurenhammer1991}. Such geometric structure provides a principled foundation for decoupling collision-free path generation from subsequent multi-contact kinematic reconstruction, and can be constructed efficiently via sweepline-based Voronoi computation \cite{Fortune1987}. To the best of the authors' knowledge, there has been no prior work that explicitly addresses motion planning for convex rolling bodies while incorporating coupled multi-contact rolling kinematics.

A key observation is that many of the above difficulties are geometric before they are kinematic: in multi-contact rolling, the feasible set is largely determined by (i) collision-free separation between rolling bodies/obstacles on the host surface and (ii) the limited accessible regions of that surface. Voronoi-based reasoning is well suited to clearance-aware planning, since the (generalized) Voronoi diagram yields a principled maximum-clearance skeleton of the free space. Paths extracted from this structure tend to stay away from obstacle boundaries and mutual-exclusion regions \cite{Choset2005,Aurenhammer1991,Choset2000IJRR}. This motivates a decoupled pipeline: first compute a collision-free route on the spherical manifold using the Voronoi structure (constructed efficiently via sweepline-style methods \cite{Fortune1987}), then reconstruct an admissible rolling motion by mapping the resulting surface path to rolling-consistent trajectories through the multi-contact kinematics. By separating where to go (clearance-aware geometric planning) from how to roll (nonholonomic multi-contact reconstruction), the coupled planning problem becomes more tractable while still respecting the rolling-contact constraints.

This work presents planning with obstacle avoidance framework for multi-contact spherical rolling systems, where multiple rolling bodies move on a shared host surface and must satisfy coupled nonholonomic rolling constraints together with collision avoidance. A clearance-oriented spherical Voronoi roadmap is constructed from near-uniform best-candidate seeds after spherical-cap feasibility filtering, and shortest paths are obtained by graph search with embedded collision checking against $3$D obstacles and mutual-exclusion regions. The resulting discrete routes are converted into differentiable, manifold-consistent trajectories via intrinsic log--exp interpolation on $\mathbb{S}^2$, and admissible multi-contact rolling motions are finally reconstructed using the generalized Montana multi-contact model and validated in forward simulation across varying numbers of sharp turns, seed densities, and computation times.

\noindent
The paper is structured as follows: Section~\ref{sec:multi_sphere_rolling} outlines the spherical multi-contact kinematic model from the Montana framework. Section~\ref{sec:voronoipathplanning} presents the Voronoi roadmap construction, graph search, smoothing, and kinematic reconstruction pipeline. Section~\ref{Sec:ResultsandDiscussion} reports simulation results and discusses the observed trade-offs and limitations.

\section{Multi-Contact Rolling Sphere Kinematic Model}
\label{sec:multi_sphere_rolling}
In this section, the general multi-contact rolling model is specialised to a primary sphere (radius $R_o$) in contact with $n$ secondary spheres (radii $R_{f,i}$), as shown in Fig.~\ref{fig:results:overview}. Montana’s contact-coordinate framework is used, and the notation from the general derivation is retained \cite{Montana1988,tafrishi2025survey}. For each contact $i$, we group the four surface coordinates into
\begin{align}
\bm{\chi}_i=\begin{bmatrix}\bm{\eta}_{o,i}^\top & \bm{\eta}_{f,i}^\top & \psi_i\end{bmatrix}^\top
\in\mathbb{R}^{5},
\bm{\eta}_{o,i}=\begin{bmatrix}u_{o,i}\\v_{o,i}\end{bmatrix},
\bm{\eta}_{f,i}=\begin{bmatrix}u_{f,i}\\v_{f,i}\end{bmatrix}.
\label{eq:eta_chi_def}
\end{align}
We then define the stacked contact state
\begin{align*}
\bm{\chi}
=
\begin{bmatrix}
\bm{\chi}_1^\top & \cdots & \bm{\chi}_n^\top
\end{bmatrix}^{\!\top}\in\mathbb{R}^{5n}.
\label{eq:chi_def}
\end{align*}
The general multi-contact kinematics are written compactly as \cite{tafrishi2025survey}
\begin{equation}
\dot{\bm{\chi}}=\bm{D}_o\bm{\nu}_o+\bm{D}_f\bm{\nu}_f,
\label{eq:multi_contact_main}
\end{equation}
where the generalized relative velocity $\bm{\nu}_o\in\mathbb{R}^{6}$ is the twist of the primary sphere,
\begin{align}
\bm{\nu}_o
=
\begin{bmatrix}
\bm{\omega}_o\\
\bm{v}_o
\end{bmatrix},
\label{eq:nu_o_def}
\end{align}
with $\bm{\omega}_o,\bm{v}_o\in\mathbb{R}^{3}$ denoting the angular and linear velocities of the primary-sphere center (in the same reference-frame convention used throughout the paper). Likewise, for each secondary sphere $i$ we define its twist
\begin{align}
\bm{\nu}_{f,i}
=
\begin{bmatrix}
\bm{\omega}_{f,i}\\
\bm{v}_{f,i}
\end{bmatrix}
\in\mathbb{R}^{6},
\label{eq:nu_fi_def}
\end{align}
and the stacked secondary-sphere velocity vector
\begin{align}
\bm{\nu}_f
=
\begin{bmatrix}
\bm{\nu}_{f,1}^\top & \cdots & \bm{\nu}_{f,n}^\top
\end{bmatrix}^{\!\top}
\in\mathbb{R}^{6n}.
\label{eq:nu_f_def}
\end{align}
With these definitions, $\bm{D}_o\in\mathbb{R}^{5n\times 6}$ captures how the primary-sphere twist drives the evolution of all contact coordinates, while $\bm{D}_f\in\mathbb{R}^{5n\times 6n}$ collects the contributions of the secondary-sphere twists (block-diagonal over contacts). Equivalently, Equation~\eqref{eq:multi_contact_main} can be written per contact as
\begin{align}
\dot{\bm{\chi}}_i
=
\bm{D}_{o,i}\bm{\nu}_o
+
\bm{D}_{f,i}\bm{\nu}_{f,i},
\qquad i=1,\dots,n,
\label{eq:multi_contact_per_contact}
\end{align}
and the global matrices $\bm{D}_o,\bm{D}_f$ are assembled from per-contact blocks as summarized in~\eqref{eq:DoDf_assembly} and \eqref{eq:contact_blocks}, see also the general construction in \cite{tafrishi2025survey}.

To define each local contact configuration explicitly, the contact point on the primary sphere at contact $i$ is
\begin{align}
\bm{c}_{o,i}
=
\begin{bmatrix}
-R_o\sin u_{o,i}\cos v_{o,i}\\
\;\;R_o\sin v_{o,i}\\
-R_o\cos u_{o,i}\cos v_{o,i}
\end{bmatrix},
\label{eq:coi}
\end{align}
and the corresponding local contact point on the secondary sphere is
\begin{align}
\bm{c}_{f,i}
=
\begin{bmatrix}
-R_{f,i}\sin u_{f,i}\cos v_{f,i}\\
\;\;R_{f,i}\sin v_{f,i}\\
-R_{f,i}\cos u_{f,i}\cos v_{f,i}
\end{bmatrix}.
\label{eq:cfi}
\end{align}
Using the outward normal $\bm{n}_{o,i}=\bm{c}_{o,i}/\|\bm{c}_{o,i}\|$, the secondary-sphere center position is
\begin{align}
\bm{p}_{f,i}=\bm{c}_{o,i}+R_{f,i}\bm{n}_{o,i}.
\label{eq:pfi}
\end{align}

For spherical surfaces, the differential-geometric terms required by the Montana model are
\begin{align}
&\bm{K}_{o,i}=
\begin{bmatrix}
\frac{1}{R_o}&0\\
0&\frac{1}{R_o}
\end{bmatrix},
\;
\bm{M}_{o,i}=
\begin{bmatrix}
R_o&0\\
0&R_o\cos v_{o,i}
\end{bmatrix}, \nonumber\\
&\bm{T}_{o,i}=
\begin{bmatrix}
0&-\tan(v_{o,i})/R_o
\end{bmatrix},
\;
\bm{T}_{f,i}=
\begin{bmatrix}
0&-\tan(v_{f,i})/R_{f,i}
\end{bmatrix},\nonumber\\
&\bm{K}_{f,i}=
\begin{bmatrix}
\frac{1}{R_{f,i}}&0\\
0&\frac{1}{R_{f,i}}
\end{bmatrix},
\;
\bm{M}_{f,i}=
\begin{bmatrix}
R_{f,i}&0\\
0&R_{f,i}\cos v_{f,i}
\end{bmatrix}.
\label{eq:KMT_spheres}
\end{align}
The relative spin rotation on the tangent plane is denoted by $\bm{R}_{\psi,i}\in\mathbb{R}^{2\times 2}$, and the spin-rotated secondary curvature is
\begin{align}
\tilde{\bm{K}}_{f,i} := \bm{R}_{\psi,i}\,\bm{K}_{f,i}\,\bm{R}_{\psi,i}^{\top},
\label{eq:Ktilde_def}
\end{align}
exactly as in the general formulation.

The full kinematic matrices for (\ref{eq:multi_contact_main}) are assembled from per-contact blocks. At the global level,
\begin{align}
\bm{D}_o=
\begin{bmatrix}
\bm{D}_{o,1}\\
\vdots\\
\bm{D}_{o,n}
\end{bmatrix},
\qquad
\bm{D}_f=\operatorname{diag}\!\left(\bm{D}_{f,1},\ldots,\bm{D}_{f,n}\right),
\label{eq:DoDf_assembly}
\end{align}
and each contact block is written as
\begin{align}
\bm{D}_{o,i}
=
\begin{bmatrix}
\bm{D}^{(1)}_{o,i} & \bm{D}^{(2)}_{o,i}\\
\bm{D}^{(3)}_{o,i} & \bm{D}^{(4)}_{o,i}\\
\bm{D}^{(5)}_{o,i} & \bm{D}^{(6)}_{o,i}
\end{bmatrix},
\bm{D}_{f,i}
=
\begin{bmatrix}
\bm{D}^{(1)}_{f,i} & \bm{D}^{(2)}_{f,i}\\
\bm{D}^{(3)}_{f,i} & \bm{D}^{(4)}_{f,i}\\
\bm{D}^{(5)}_{f,i} & \bm{D}^{(6)}_{f,i}
\end{bmatrix}.
\label{eq:contact_blocks}
\end{align}

\begin{figure*}[t]
    \centering    
    \includegraphics[width=0.27\linewidth, trim=3cm 0.5cm 3cm 0.5cm, clip]{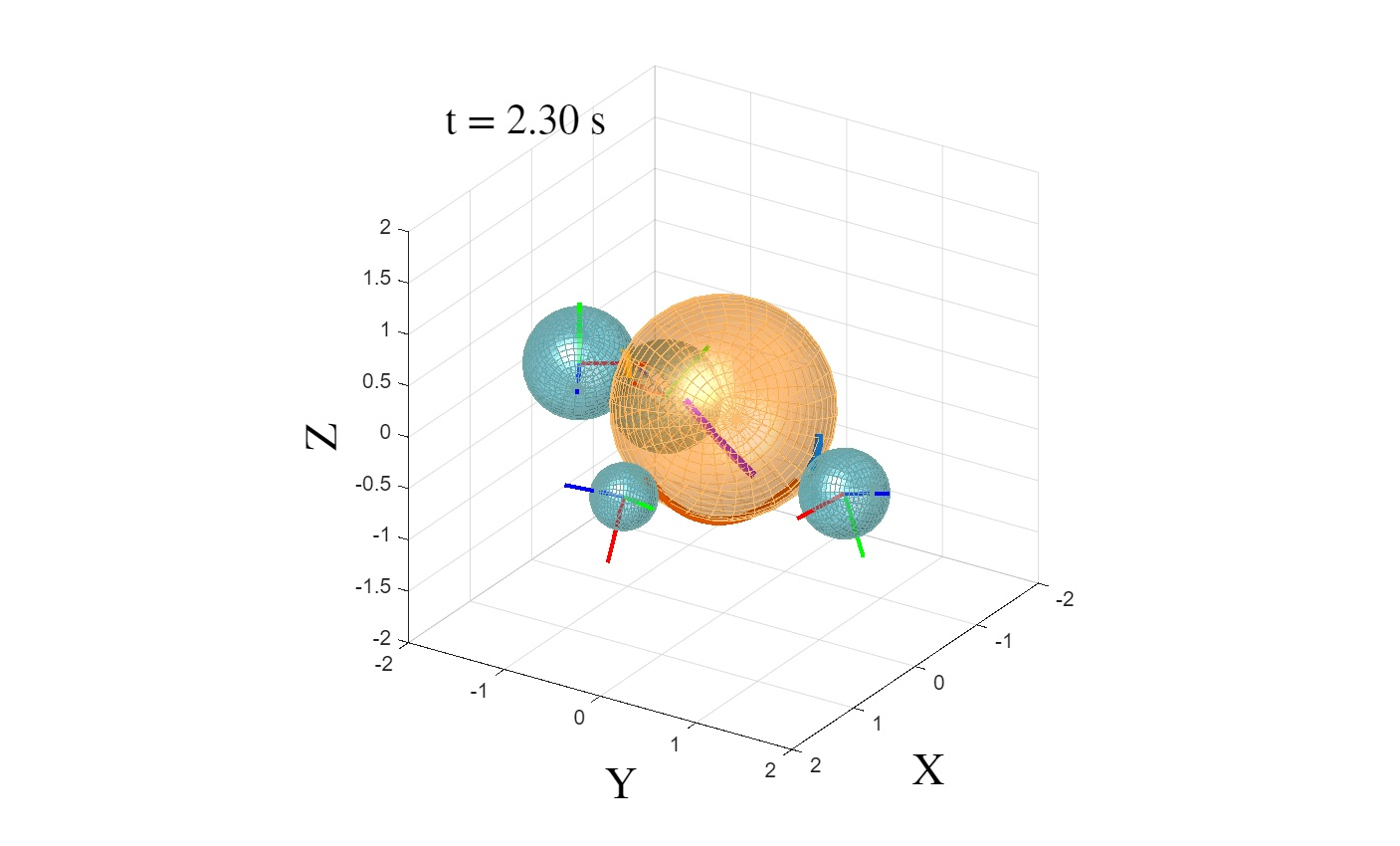}%
    \hspace{-5mm}%
    \includegraphics[width=0.27\linewidth, trim=3cm 0.5cm 3cm 0.5cm, clip]{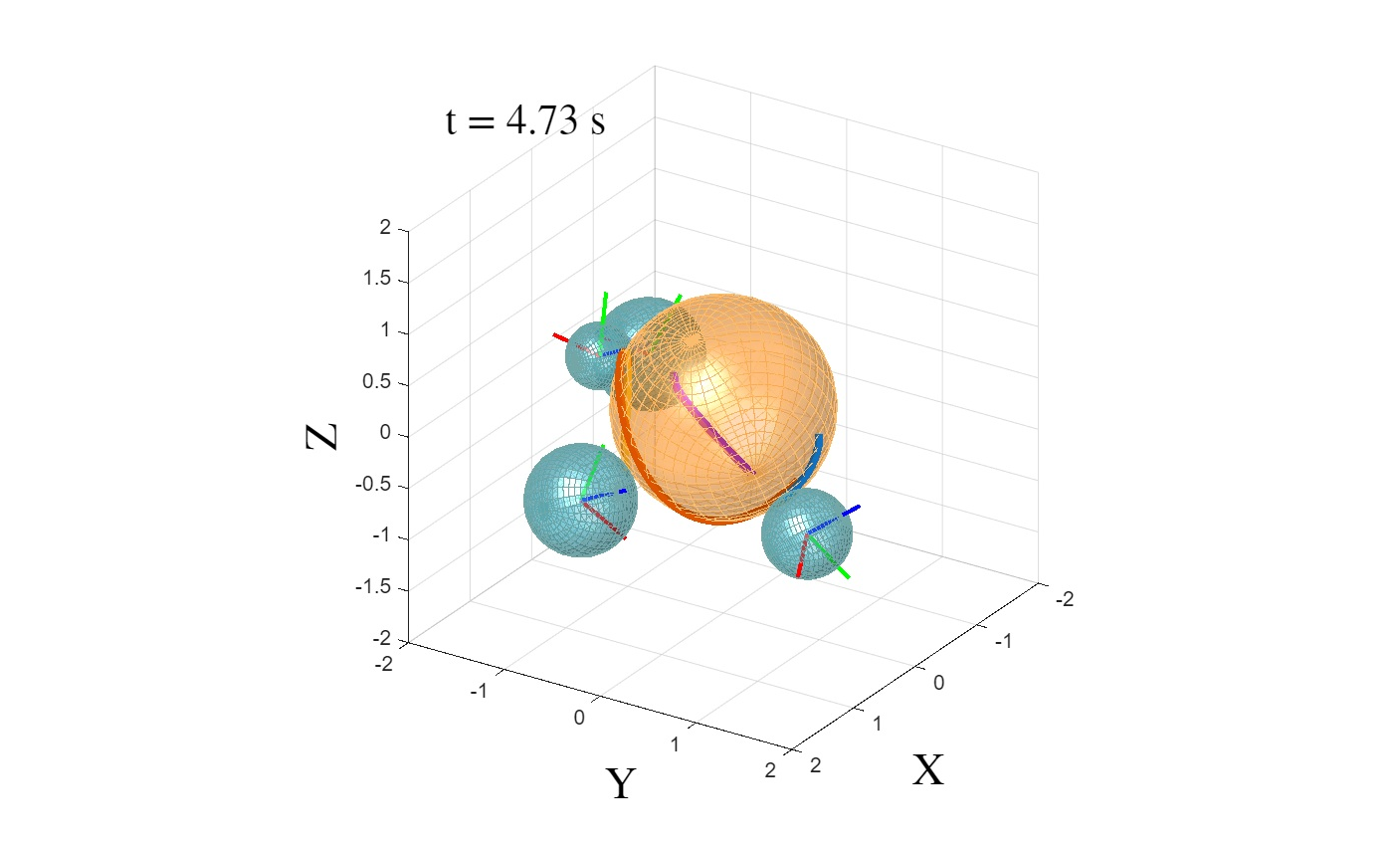}%
    \hspace{-5mm}%
    \includegraphics[width=0.27\linewidth, trim=3cm 0.5cm 3cm 0.5cm, clip]{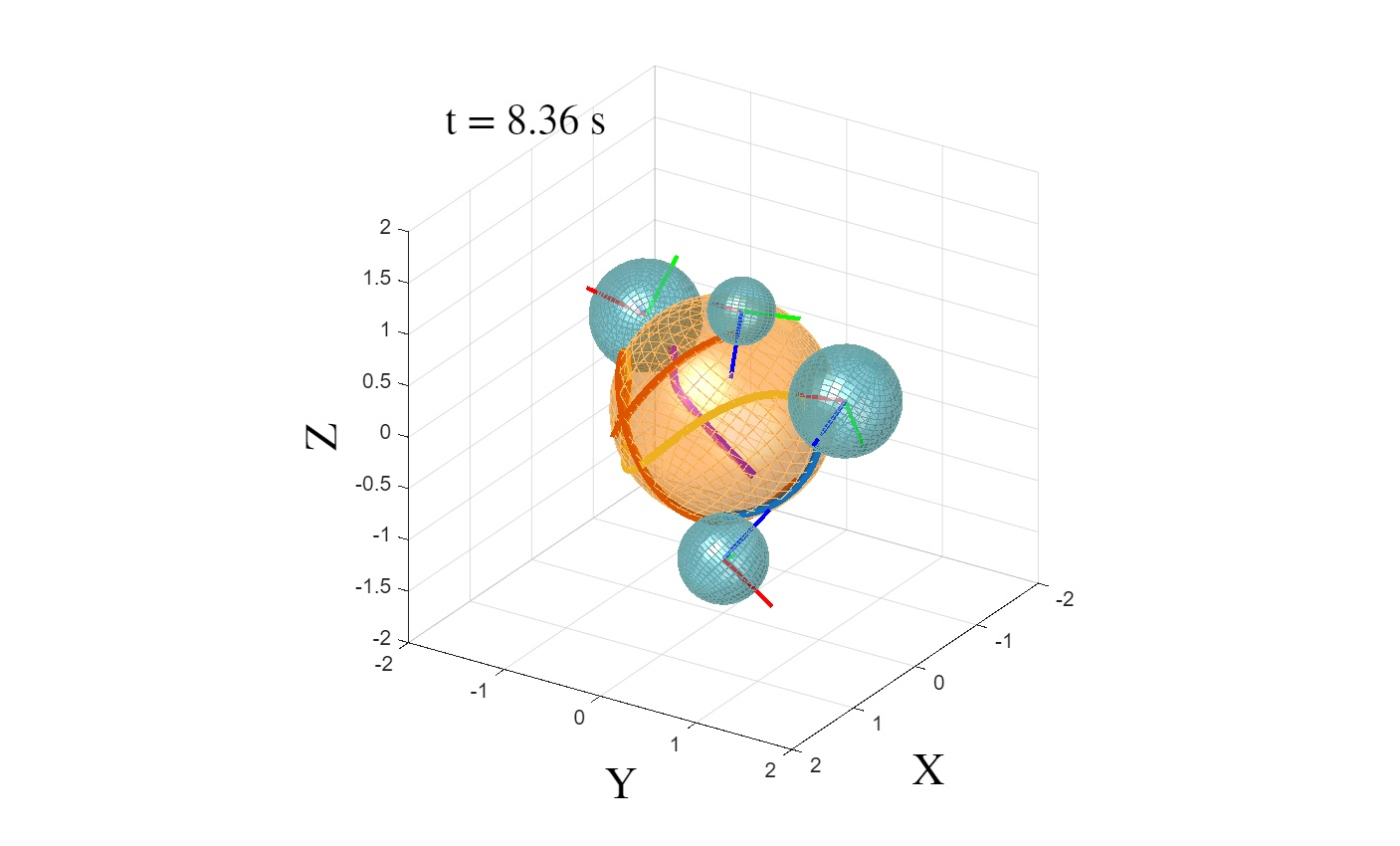}%
    \hspace{-5mm}%
    \includegraphics[width=0.27\linewidth, trim=3cm 0.5cm 3cm 0.5cm, clip]{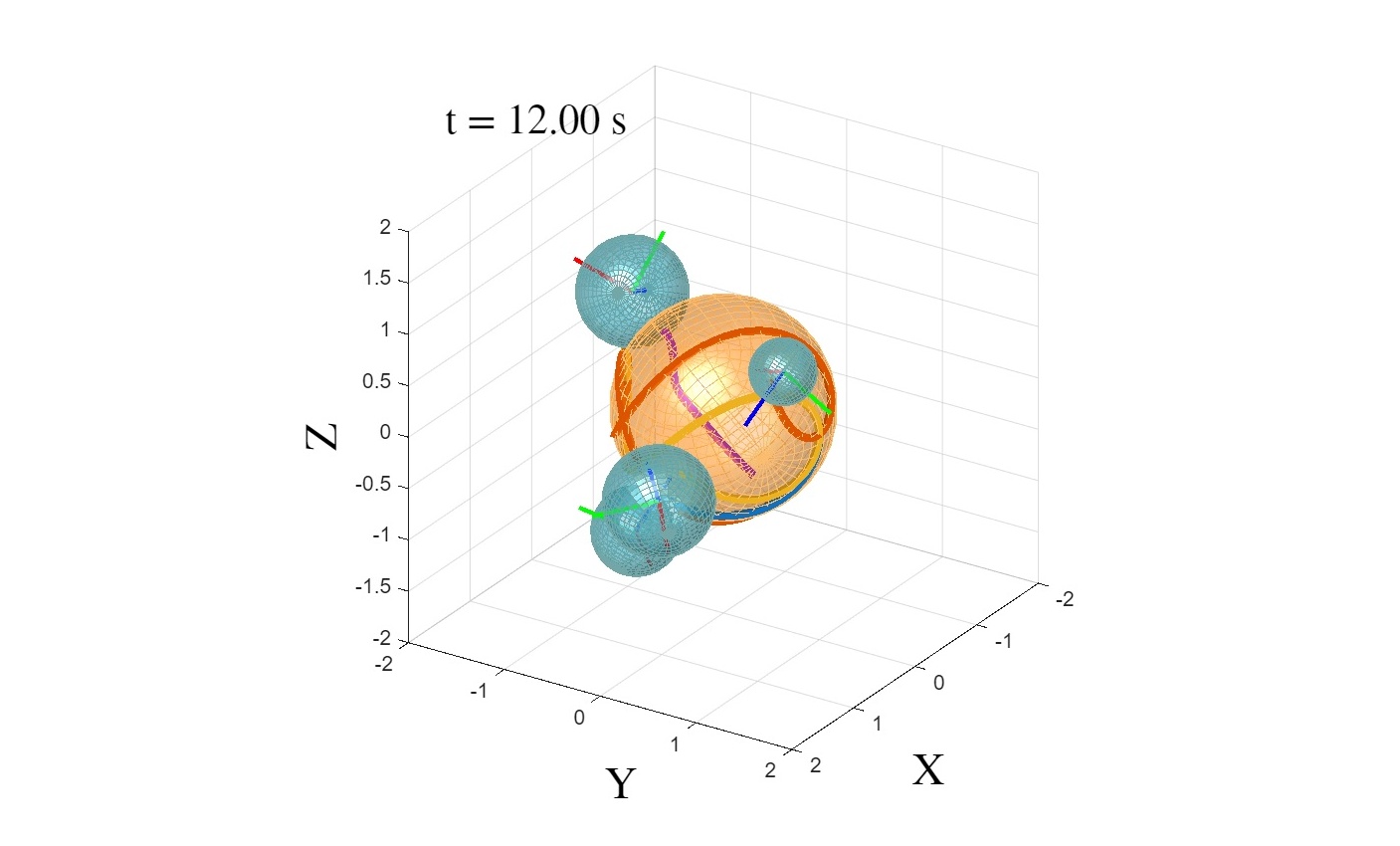} 
    \caption{
    Time evolution of the rolling-contact kinematics on the spherical manifold 
    at $t = 2.30\,\mathrm{s}$, $t = 4.73\,\mathrm{s}$, $t = 8.36\,\mathrm{s}$, and $t = 12.00\,\mathrm{s}$. 
    }   
    \label{fig:rolling_evolution}
\end{figure*}
Also, please note that $\bm{E}_1$, $\bm{E}_2$, and $\bm{E}_3$ are the constant selection matrices
$$
\bm{E}_1=\begin{bmatrix}0&0&1\end{bmatrix},\quad
\bm{E}_2=\begin{bmatrix}0&1&0\\-1&0&0\end{bmatrix},\quad
\bm{E}_3=\begin{bmatrix}1&0&0\\0&1&0\end{bmatrix}.
$$

The primary-sphere contribution at each contact is
\begin{align}
\bm{D}^{(1)}_{o,i}&=-\bm{\Upsilon}_{o,i}\tilde{\bm{K}}_{f,i}\bm{E}_3\bm{R}'_{o,i},\nonumber\\
\bm{D}^{(2)}_{o,i}&=\bm{\Upsilon}_{o,i}\!\left(\bm{E}_2\bm{H}_{o,i}+\tilde{\bm{K}}_{f,i}\bm{E}_3\bm{H}'_{o,i}\right),\nonumber\\
\bm{D}^{(3)}_{o,i}&=\bm{\Upsilon}_{f,i}\bm{K}_{o,i}\bm{E}_3\bm{R}'_{o,i},\nonumber\\
\bm{D}^{(4)}_{o,i}&=\bm{\Upsilon}_{f,i}\!\left(\bm{E}_2\bm{H}_{o,i}-\bm{K}_{o,i}\bm{E}_3\bm{H}'_{o,i}\right),\nonumber\\
\bm{D}^{(5)}_{o,i}&=\bm{T}_{f,i}\bm{M}_{f,i}\bm{D}^{(3)}_{o,i}
+\bm{T}_{o,i}\bm{M}_{o,i}\bm{D}^{(1)}_{o,i},\nonumber\\
\bm{D}^{(6)}_{o,i}&=\bm{T}_{f,i}\bm{M}_{f,i}\bm{D}^{(4)}_{o,i}
+\bm{T}_{o,i}\bm{M}_{o,i}\bm{D}^{(2)}_{o,i}
-\bm{E}_1\bm{H}_{o,i}.
\label{eq:Do_components}
\end{align}

The secondary-sphere contribution at each contact is
\begin{align}
\bm{D}^{(1)}_{f,i}&=\bm{\Upsilon}_{o,i}\tilde{\bm{K}}_{f,i}\bm{E}_3\bm{R}'_{f,i},\nonumber\\
\bm{D}^{(2)}_{f,i}&=-\bm{\Upsilon}_{o,i}\!\left(\bm{E}_2\bm{H}_{f,i}+\tilde{\bm{K}}_{f,i}\bm{E}_3\bm{H}'_{f,i}\right),\nonumber\\
\bm{D}^{(3)}_{f,i}&=-\bm{\Upsilon}_{f,i}\bm{K}_{o,i}\bm{E}_3\bm{R}'_{f,i},\nonumber\\
\bm{D}^{(4)}_{f,i}&=-\bm{\Upsilon}_{f,i}\!\left(\bm{E}_2\bm{H}_{f,i}-\bm{K}_{o,i}\bm{E}_3\bm{H}'_{f,i}\right),\nonumber\\
\bm{D}^{(5)}_{f,i}&=\bm{T}_{f,i}\bm{M}_{f,i}\bm{D}^{(3)}_{f,i}
+\bm{T}_{o,i}\bm{M}_{o,i}\bm{D}^{(1)}_{f,i},\nonumber\\
\bm{D}^{(6)}_{f,i}&=\bm{T}_{f,i}\bm{M}_{f,i}\bm{D}^{(4)}_{f,i}
+\bm{T}_{o,i}\bm{M}_{o,i}\bm{D}^{(2)}_{f,i}
+\bm{E}_1\bm{H}_{f,i}.
\label{eq:Df_components}
\end{align}
All auxiliary terms $\bm{R}'_{\{\cdot\}}$, $\bm{H}_{\{\cdot\}}$, and $\bm{H}'_{\{\cdot\}}$ follow directly from the general multi-contact construction in \cite{TafrishiIJRRMultiContact} and are used here without modification. Also, the coupling multipliers are
\begin{align}
\bm{\Upsilon}_{o,i}
&=\bm{M}_{o,i}^{-1}\!\left(\bm{K}_{o,i}+\tilde{\bm{K}}_{f,i}\right)^{-1},\nonumber\\
\bm{\Upsilon}_{f,i}
&=\bm{M}_{f,i}^{-1}\bm{R}_{\psi,i}\!\left(\bm{K}_{o,i}+\tilde{\bm{K}}_{f,i}\right)^{-1}
\label{eq:Upsilon_defs}
\end{align}
which reduce to the closed spherical forms reported in the main derivation.

The equations developed in \eqref{eq:multi_contact_main}--\eqref{eq:DoDf_assembly} establish a generalized kinematic model for multi-contact rolling bodies, with direct applicability to reconfigurable rolling robots and rolling-contact dexterous manipulation. To verify the consistency of the derived model, a representative simulation is conducted for four simultaneous contacts, where secondary spheres undergo spin-rolling motion around the fixed primary sphere by numerically integrating~\eqref{eq:multi_contact_main} with the assembled model in~\eqref{eq:DoDf_assembly}. The resulting time evolution of the contact states is illustrated in Fig.~\ref{fig:rolling_evolution}, confirming that the coupled no-slip constraints are maintained throughout the motion. Such multi-contact kinematic behaviour is fundamental to applications such as reconfigurable rolling robots, where multiple modules traverse a host sphere, and rolling-contact dexterous manipulation, where fingertip rolling enables stable object reorientation without releasing the grasp.

\section{Voronoi-Based Path Planning}
\label{sec:voronoipathplanning}
\begin{figure}[t!]
	\centering
	\includegraphics[width=\columnwidth]{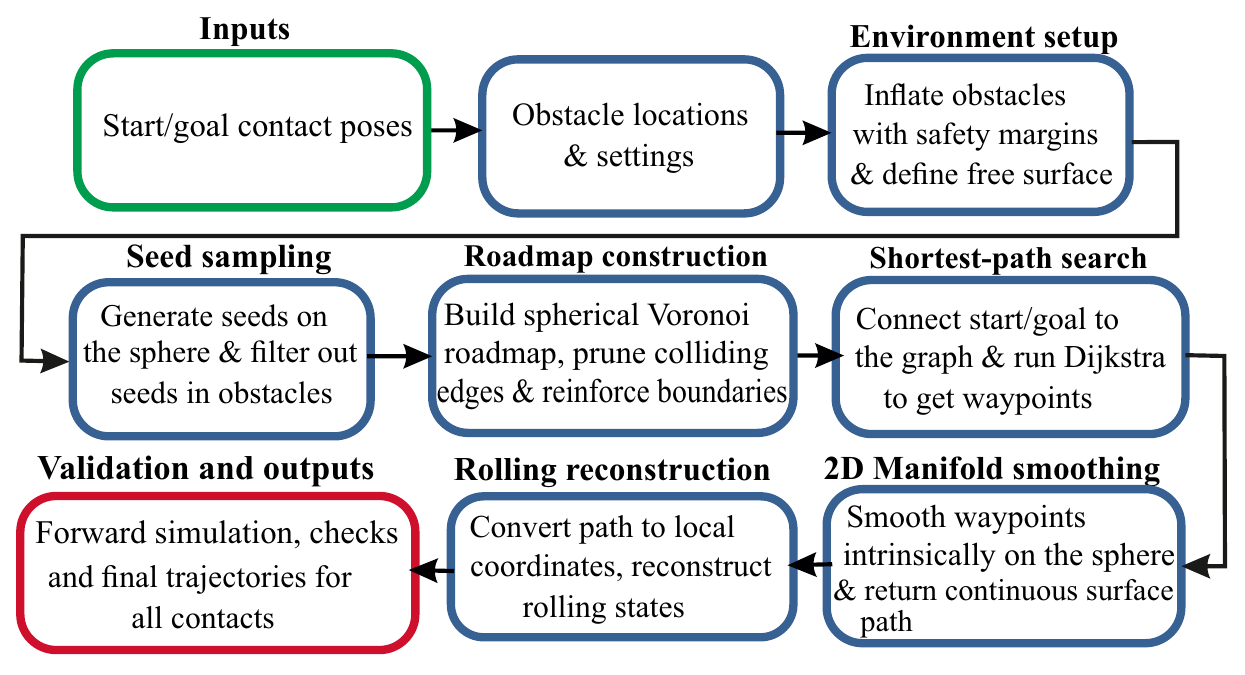}
	\caption{Compact overview of the spherical Voronoi planning and rolling-motion reconstruction pipeline.}
 \label{Fig:Qing_Blockdiagram}
\end{figure}

This section presents a Voronoi-based planning and execution framework on the spherical manifold, as summarised in the block diagram of Fig.\ref{Fig:Qing_Blockdiagram}. The goal is to compute a collision-free, kinematically feasible trajectory on \(S^2\) (in its Cartesian embedding) and then map this Cartesian trajectory to the 2D Montana contact-coordinate manifold for rolling execution. 

\subsection{Problem Statement}
\label{statement}

\begin{figure}[t!]
	\centering
	\includegraphics[width=2.4 in]{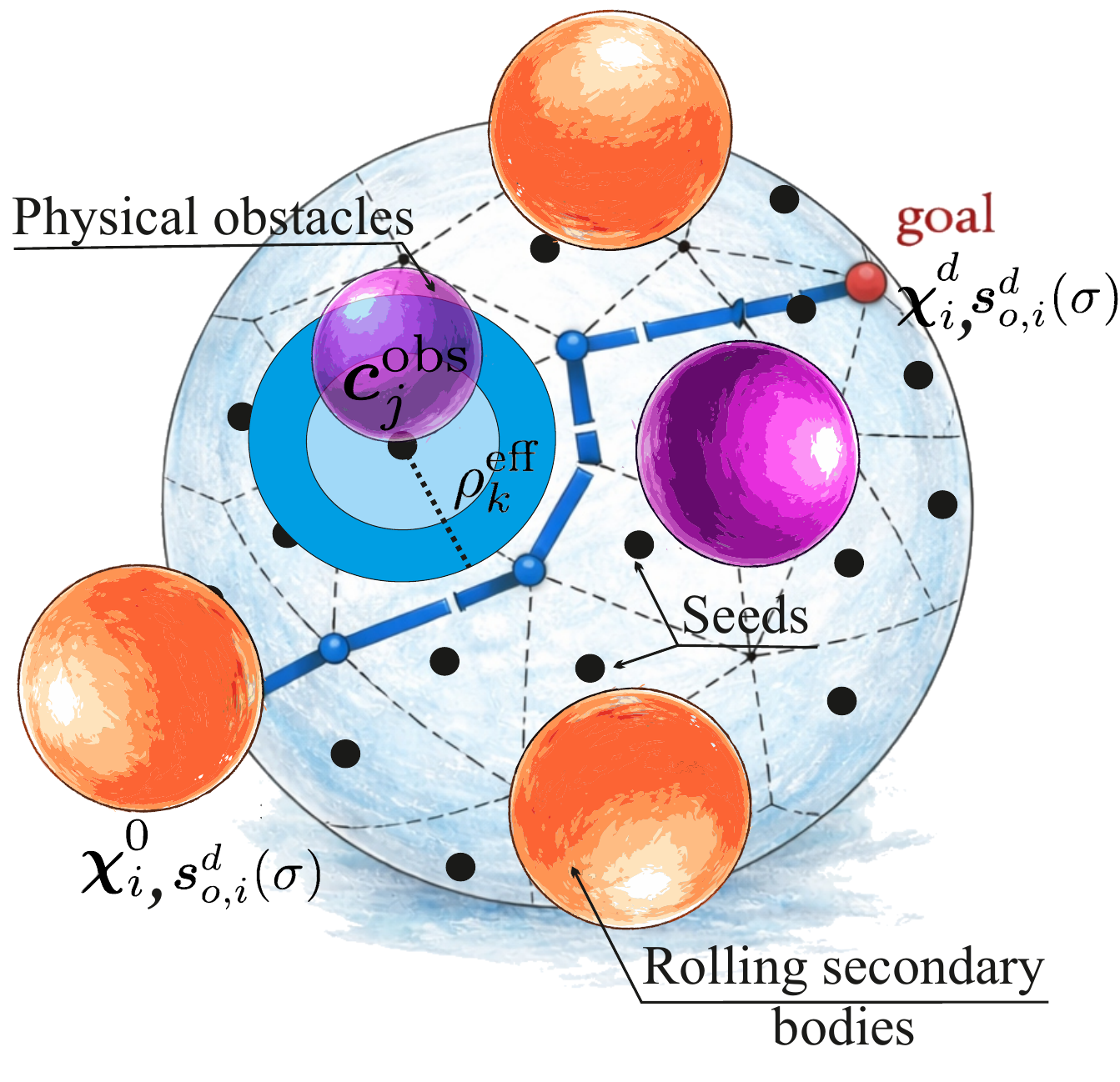}
	\caption{Voronoi-roadmap planning on the spherical 2D contact manifold.}
 \label{Fig:PlanningStatement}
\end{figure}

We consider a multi-contact spherical rolling system composed of one \emph{primary} sphere of radius $R_o$ and $n$ \emph{secondary} bodies, specialised here to secondary spheres of radii $R_{f,i}$, $i=1,\dots,n$. At contact $i$, the local Montana contact state is $\bm{\chi}_i$
where $(u_{o,i},v_{o,i})$ and $(u_{f,i},v_{f,i})$ are the local surface coordinates on the primary and $i$-th secondary sphere, and $\psi_i$ is the relative spin angle at the contact. The stacked state is $\bm{\chi}=\mathrm{col}(\bm{\chi}_1,\dots,\bm{\chi}_n)\in\mathbb{R}^{5n}$ and its evolution is governed by the multi-contact Montana kinematics in \eqref{eq:multi_contact_main}. In this section, the primary sphere is fixed at the workspace center and undergoes pure rotation (i.e., $\bm{v}_o=\bm{0}$ in~\eqref{eq:nu_o_def}), while each secondary sphere remains in continuous no-slip rolling contact with the primary surface as shown in Fig. \ref{Fig:PlanningStatement}. 

We assume stationary obstacles induce forbidden regions on the primary sphere surface. These are represented as spherical caps
$$
\mathcal{O}=\{(\bm{c}^{\mathrm{obs}}_j,\rho_j)\}_{j=1}^{J},
$$
where $\bm{c}^{\mathrm{obs}}_j\in\mathbb{S}^2$ is the cap-center direction, $\rho_j>0$ is the angular radius (radians), and $J$ is the number of caps. With a safety margin $\delta\ge0$, the admissible region on the unit sphere is
$$
\mathcal{F}:=\Big\{\bm{s}\in\mathbb{S}^2\ \Big|\ d_{\mathbb{S}^2}(\bm{s},\bm{c}^{\mathrm{obs}}_j)\ge \rho_j+\delta,\ \forall j\Big\},
$$
where $\mathbb{S}^2:=\{\bm{x}\in\mathbb{R}^3\mid \|\bm{x}\|=1\}$ and $d_{\mathbb{S}^2}(\cdot,\cdot)$ is the spherical geodesic distance (defined consistently with the Voronoi construction in this section). Using the Cartesian embedding induced by the primary contact map \eqref{eq:coi}, we represent each primary-side contact location by the corresponding unit direction
$$
\bm{s}_{o,i}:=\frac{1}{R_o}\,\bm{c}_{o,i}(u_{o,i},v_{o,i})\in\mathbb{S}^2.
$$

The planning objective is to compute collision-free, piece-wise smooth surface trajectories for the primary-side contact directions,
$$
\bm{s}^{d}_{o,i}(\sigma)\in\mathcal{F},\qquad \sigma\in[0,1],\qquad i=1,\dots,n,
$$
that connect prescribed initial and goal contact configurations. Equivalently, this defines trajectories in local coordinates $(u^d_{o,i}(\sigma),v^d_{o,i}(\sigma))$ through $\bm{s}^{d}_{o,i}(\sigma)=\bm{c}_{o,i}(u^d_{o,i}(\sigma),v^d_{o,i}(\sigma))/R_o$. The initial states are
$$
\bm{\chi}^0_i =
\begin{bmatrix}
u^0_{o,i} & v^0_{o,i} & u^0_{f,i} & v^0_{f,i} & \psi^0_i
\end{bmatrix}^{\!\top},
\qquad i=1,\dots,n,
$$
with the corresponding initial directions $\bm{s}^0_{o,i}$ obtained from the above embedding; goal states are defined analogously. Consistent with many rolling-contact mechanisms (e.g., reconfigurable rolling robots and rolling-fingertip systems), we do not prescribe independent terminal trajectories for the secondary-side variables $(u_{f,i},v_{f,i},\psi_i)$; instead, these states are treated as induced variables and are reconstructed from the rolling kinematics once $\bm{s}^{d}_{o,i}(\sigma)$ is obtained. The generated trajectories are required to be at least $C^{1}$ (and $C^{2}$ after smoothing) to support stable numerical integration and robust kinematic reconstruction.

\subsection{Voronoi-Graph-Based Planner}
\label{subsec:voronoi_search}

With the planning objective defined, we now construct the roadmap on which paths will be searched. We first require a set of approximately uniform seed directions on the unit sphere to initialize the spherical Voronoi roadmap. To ensure that sampling respects the intrinsic geometry of the configuration space, seeds are generated directly on the embedded manifold $\mathbb{S}^2\subset\mathbb{R}^3$, where $\mathbb{S}^2:=\{\bm{x}\in\mathbb{R}^3\mid \|\bm{x}\|=1\}$. Random vectors in $\mathbb{R}^3$ are projected onto $\mathbb{S}^2$ by normalization. Since naive normalization alone can lead to uneven coverage, we employ a best-candidate (max--min) strategy that maximizes angular separation between samples, producing a blue-noise-like distribution on $\mathbb{S}^2$ \cite{Muller1959,Marsaglia1972,Bridson2007}. Let $\{\bm{q}_1,\dots,\bm{q}_{j-1}\}$ denote the accepted seeds and let $\bm{r}$ be a candidate direction on $\mathbb{S}^2$. Using the spherical (geodesic) distance
\begin{align}
&d_{\mathbb{S}^2}(\bm{x},\bm{y})
=
\arccos\!\big(\mathrm{clip}(\bm{x}^\top \bm{y})\big),\nonumber\\
&\mathrm{clip}(z):=\min(\max(z,-1),1),
\label{eq:geo_dist_voronoi}
\end{align}
the next seed is selected as
\begin{equation*}
\bm{q}_j
=
\arg\max_{\bm{r}\in\mathbb{S}^2}
\left\{
\min_{i<j} d_{\mathbb{S}^2}(\bm{r},\bm{q}_i)
\right\}.
\label{eq:best_candidate_voronoi}
\end{equation*}

To account for obstacle feasibility on the host sphere, we use the spherical-cap model introduced in Section~III.A. For an obstacle cap with center direction $\bm{c}^{\mathrm{obs}}_k$ and angular radius $\rho_k$ (radians), we inflate the forbidden region by a safety margin $\epsilon\ge 0$ and an additional angular footprint $\alpha_{\mathrm{sec}}\ge 0$ that captures the secondary-body clearance on the sphere. For spherical modules this footprint can be chosen as $\alpha_{\mathrm{sec}}=\arcsin(R_{f}/R_o)$ (or $\alpha_{\mathrm{sec}}\approx R_f/R_o$ for small $R_f/R_o$). The effective angular radius is
\begin{equation}
\rho_k^{\mathrm{eff}}=\rho_k+\alpha_{\mathrm{sec}}+\epsilon,
\label{eq:rho_eff_voronoi}
\end{equation}
and a seed $\bm{q}$ is retained if it satisfies
\begin{equation}
d_{\mathbb{S}^2}(\bm{q},\bm{c}^{\mathrm{obs}}_k)\ \ge\ \rho_k^{\mathrm{eff}}
\qquad \text{for all } k.
\label{eq:seed_feasible_voronoi}
\end{equation}
The resulting collision-free seeds form the basis for roadmap construction.

Given the filtered seed set $\{\bm{s}_i\}_{i=1}^{n_s}\subset\mathbb{S}^2$, a spherical Voronoi tessellation is constructed. The Voronoi vertices
\begin{equation*}
V_{\mathrm{vor}}=\{\bm{v}_1,\dots,\bm{v}_{n_v}\}\subset \mathbb{S}^2
\label{eq:V_vor}
\end{equation*}
approximate medial-axis locations of the free manifold. Adjacency relations between Voronoi cells induce an initial edge set $E_{\mathrm{vor}}$, with geodesic edge weights
\begin{equation*}
w_{ij}=d_{\mathbb{S}^2}(\bm{v}_i,\bm{v}_j)
=
\arccos\!\big(\mathrm{clip}(\bm{v}_i^\top \bm{v}_j)\big).
\label{eq:edge_weight_voronoi}
\end{equation*}
While this structure captures intrinsic geodesic connectivity in open regions, Voronoi degeneracies may arise near inflated obstacle caps, leading to reduced vertex density and weakened boundary representation.

To stabilize roadmap topology near obstacle boundaries, we introduce a structured boundary embedding mechanism. For each inflated spherical cap centered at $\bm{c}^{\mathrm{obs}}_k$, a regular spherical hexagon is constructed as a circumscribed boundary structure. Let
\begin{equation*}
\mathcal{H}_k=\{\bm{h}_{k,1},\dots,\bm{h}_{k,6}\}
\label{eq:hex_set}
\end{equation*}
denote the hexagonal boundary vertices. Consecutive vertices are connected by collision-free geodesic arcs to form a closed loop, generating a boundary edge set $E_{\mathrm{hex}}$. This deterministic boundary insertion compensates for Voronoi sparsity and enforces a uniformly spaced representation around obstacle caps. The hexagonal boundary vertices are placed at angular 
radius
\begin{equation}
    \delta_{\mathrm{hex},k} = \frac{\rho_k^{\mathrm{eff}}}{\cos(\pi/6)},
    \label{eq:delta_hex}
\end{equation}
from the cap center $\bm{c}_k^{\mathrm{obs}}$, so that the spherical hexagon circumscribes. To further enhance robustness, all hexagonal boundary vertices within the same cap are mutually connected via 
collision-free geodesic edges: for any pair $\bm{p},\bm{q}\in\mathcal{H}_k$, an additional bridging 
edge is introduced if
\begin{equation}
d_{\mathbb{S}^2}(\bm{p},\bm{q})\leq 2\delta_{\mathrm{hex},k},
\label{eq:bridge_condition}
\end{equation}
and the connecting geodesic segment is collision-free, 
thereby forming a complete local graph over all six 
boundary vertices of each obstacle cap. The resulting edge set $E_{\mathrm{bridge}}$ effectively constructs a local $\varepsilon$-graph over boundary vertices, mitigating connectivity loss in narrow passages and preventing graph fragmentation. The complete edge set is defined as
\begin{equation*}
E
=
E_{\mathrm{vor}}
\cup
E_{\mathrm{hex}}
\cup
E_{\mathrm{bridge}}.
\label{eq:edge_union}
\end{equation*}
All edge endpoints are collected and merged using tolerance-based clustering to form the final node set $V$, yielding an undirected weighted graph
\begin{equation*}
G=(V,E).
\label{eq:graph_def}
\end{equation*}
We assume start and goal configurations on the sphere, $\bm{s}^0,\bm{s}^g\in\mathbb{S}^2$, are connected to nearby vertices via collision-free geodesic edges. In our planner, a weighted shortest path is then computed using modified Dijkstra's algorithm on spherical convex manifold \cite{Dijkstra1959,singamaneni2024survey}; Hence, let
\begin{equation*}
\pi=\{i_0,i_1,\dots,i_K\}
\label{eq:path_index_seq}
\end{equation*}
denote the resulting vertex sequence, and define the corresponding discrete spherical trajectory as
\begin{equation*}
\bm{P}_d=\{\bm{v}_{i_0},\dots,\bm{v}_{i_K}\}.
\label{eq:discrete_traj}
\end{equation*}
The discrete path obtained from the Voronoi graph search is thus a sequence of points on $\mathbb{S}^2$. Although collision-free, the piecewise geodesic connection between successive nodes may introduce non-smooth curvature variations.

To ensure geometric consistency and differentiability on the spherical manifold, we perform a Riemannian log--exp smoothing procedure in Fig.\ref{Fig:Qing_Blockdiagram}. Let $\bm{c}_{\mathrm{ref}}\in\mathbb{S}^2$ denote a reference direction chosen as the normalized mean direction of the discrete path samples, and let $\{\bm{e}_1,\bm{e}_2\}$ be an orthonormal basis of the tangent plane $T_{\bm{c}_{\mathrm{ref}}}\mathbb{S}^2$. Each spherical point $\bm{p}_i\in\bm{P}_d$ is mapped to the tangent plane via the logarithmic map:
\begin{equation*}
\theta_i=\arccos(\bm{c}_{\mathrm{ref}}^\top \bm{p}_i),
\label{eq:theta_log}
\end{equation*}
\begin{equation*}
\bm{\zeta}_i
=
\frac{\theta_i}{\sin\theta_i}
\left(
\bm{p}_i-(\bm{c}_{\mathrm{ref}}^\top \bm{p}_i)\bm{c}_{\mathrm{ref}}
\right),
\label{eq:log_map}
\end{equation*}
and expressed in local tangent coordinates $(\xi_i,\eta_i)$ with respect to $\{\bm{e}_1,\bm{e}_2\}$ (to avoid overloading the contact coordinates $u_{(\cdot)},v_{(\cdot)}$ used in Section~II). Cubic interpolation is then applied independently to $\xi$ and $\eta$ to obtain a $C^1$ continuous trajectory in the tangent space. The smoothed points are mapped back to the sphere using the exponential map \cite{smoothNoakes1989}:
\begin{equation*}
\bm{p}(\sigma)
=
\cos\|\bm{\zeta}(\sigma)\|\, \bm{c}_{\mathrm{ref}}
+
\sin\|\bm{\zeta}(\sigma)\|\,
\frac{\bm{\zeta}(\sigma)}{\|\bm{\zeta}(\sigma)\|},
\label{eq:exp_map}
\end{equation*}
where $\sigma$ is the path parameter. This log--exp interpolation ensures that smoothing is performed intrinsically on the manifold, preserving the spherical constraint while eliminating curvature discontinuities.The resulting $C^2$ trajectory is passed to the inverse kinematics stage described next..

\subsection{Inverse Kinematics Based on the Montana Framework}
\label{subsec:inverse_kinematics_montana}

This section provides a compact inverse 
map for the planner: we plan the primary-side surface
trajectory $\bm{\eta}_{o,i}(t) = [u_{o,i}(t),\, v_{o,i}(t)]^\top$
and reconstruct the remaining contact states $\bm{\eta}_{f,i}(t)$
and $\psi_i(t)$ under the no-slip (pure rolling) assumption.
The starting point is the multi-contact kinematic model in
Section~\ref{sec:multi_sphere_rolling}, where each contact
satisfies $\dot{\bm{\chi}}_i = \bm{D}_{o,i}\bm{\nu}_o +
\bm{D}_{f,i}\bm{\nu}_{f,i}$ with
$\bm{\chi}_i = [\bm{\eta}_{o,i}^\top\; \bm{\eta}_{f,i}^\top\;
\psi_i]^\top$.

For a single contact $i$, the Montana contact-coordinate
form can be written in a decoupled structure with the relative
twist expressed in the local contact frame. The normal spin
$\omega_{z,i}$ enters as a free input that does not induce
tangential slip, and is either prescribed by a controller or
set to zero when only in-surface rolling is required.
Under the pure rolling assumption with $\omega_{z,i}$ treated
as above, denote the relative angular velocity by
$\bm{\omega}_{rel,i} = [\omega_{x,i},\, \omega_{y,i},\,
\omega_{z,i}]^\top$ and let $\bm{\omega}_{xy,i} =
[\omega_{x,i},\, \omega_{y,i}]^\top$. The contact-state
rates depend only on $\bm{\omega}_{rel,i}$ and, for spherical
surfaces, the primary-side rates admit the compact map
\begin{equation*}
    \dot{\bm{\eta}}_{o,i} = \bm{g}_{o,i}(\bm{\chi}_i)\,
    \bm{\omega}_{xy,i},
    \label{eq:primary_map}
\end{equation*}
where $\bm{g}_{o,i}(\bm{\chi}_i) \in \mathbb{R}^{2\times2}$
is the primary-side angular Jacobian induced by the Montana
operators (equivalently obtained from the
$(\bm{K}_{o,i} + \tilde{\bm{K}}_{f,i})^{-1}$ and
$\bm{M}_{o,i}^{-1}$ structure in
Section~\ref{sec:multi_sphere_rolling}), and is generically
full rank for spherical contacts away from kinematic
singularities. Therefore, the planner-provided rate
$\dot{\bm{\eta}}_{o,i}$ yields the required transformed
rolling input
\begin{equation}
    \bm{\omega}_{xy,i} = \bm{g}_{o,i}^{-1}(\bm{\chi}_i)\
    \dot{\bm{\eta}}_{o,i},
    \label{eq:omega_xy_inv}
\end{equation}

\begin{figure*}[t]
    \centering
    \begin{subfigure}[t]{0.4\textwidth}
        \centering
        \includegraphics[width=\linewidth]{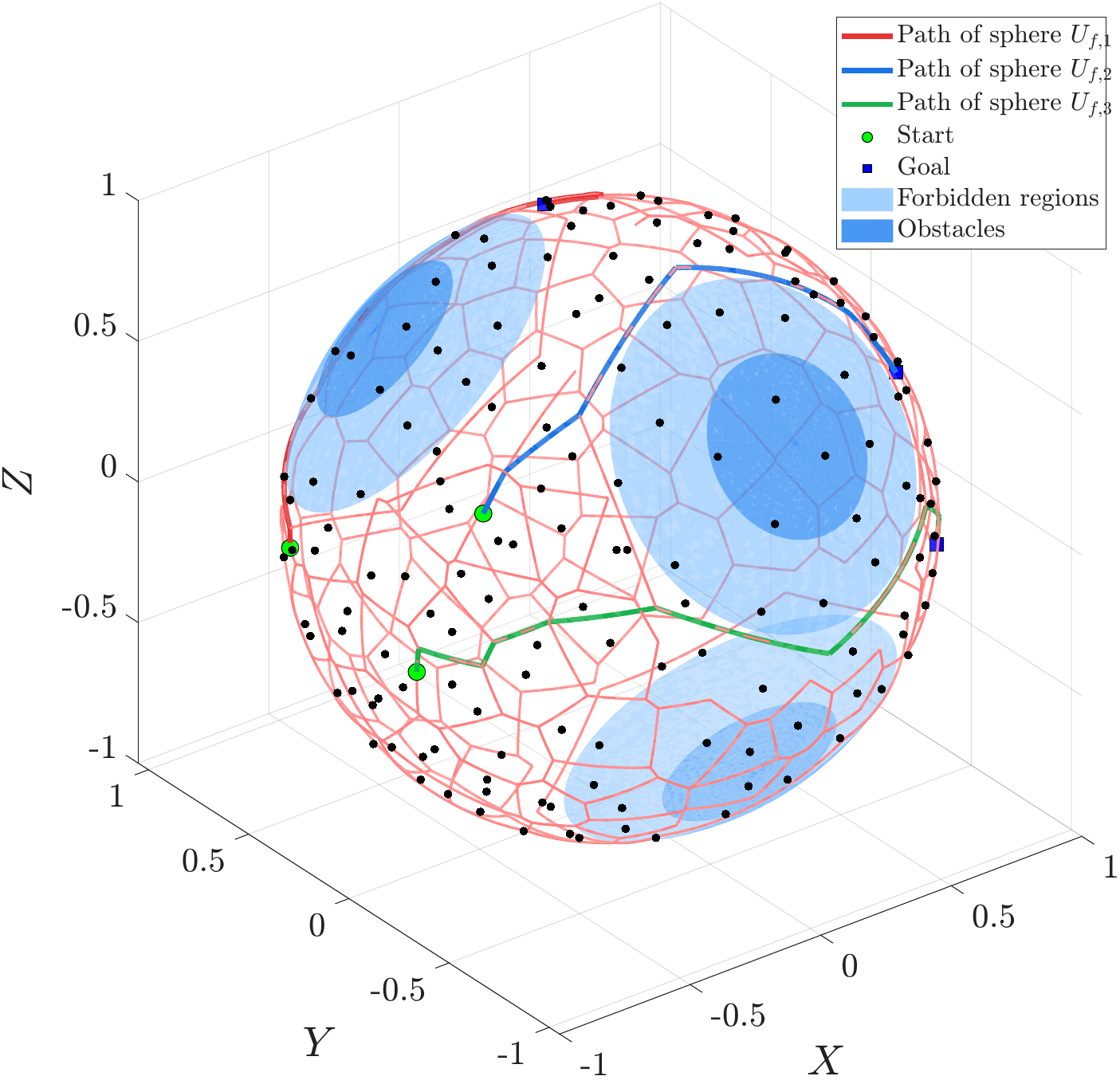}
        \caption{}\label{fig:res:a}
    \end{subfigure}
    \begin{subfigure}[t]{0.4\textwidth}
        \centering
        \includegraphics[width=3.8 in]{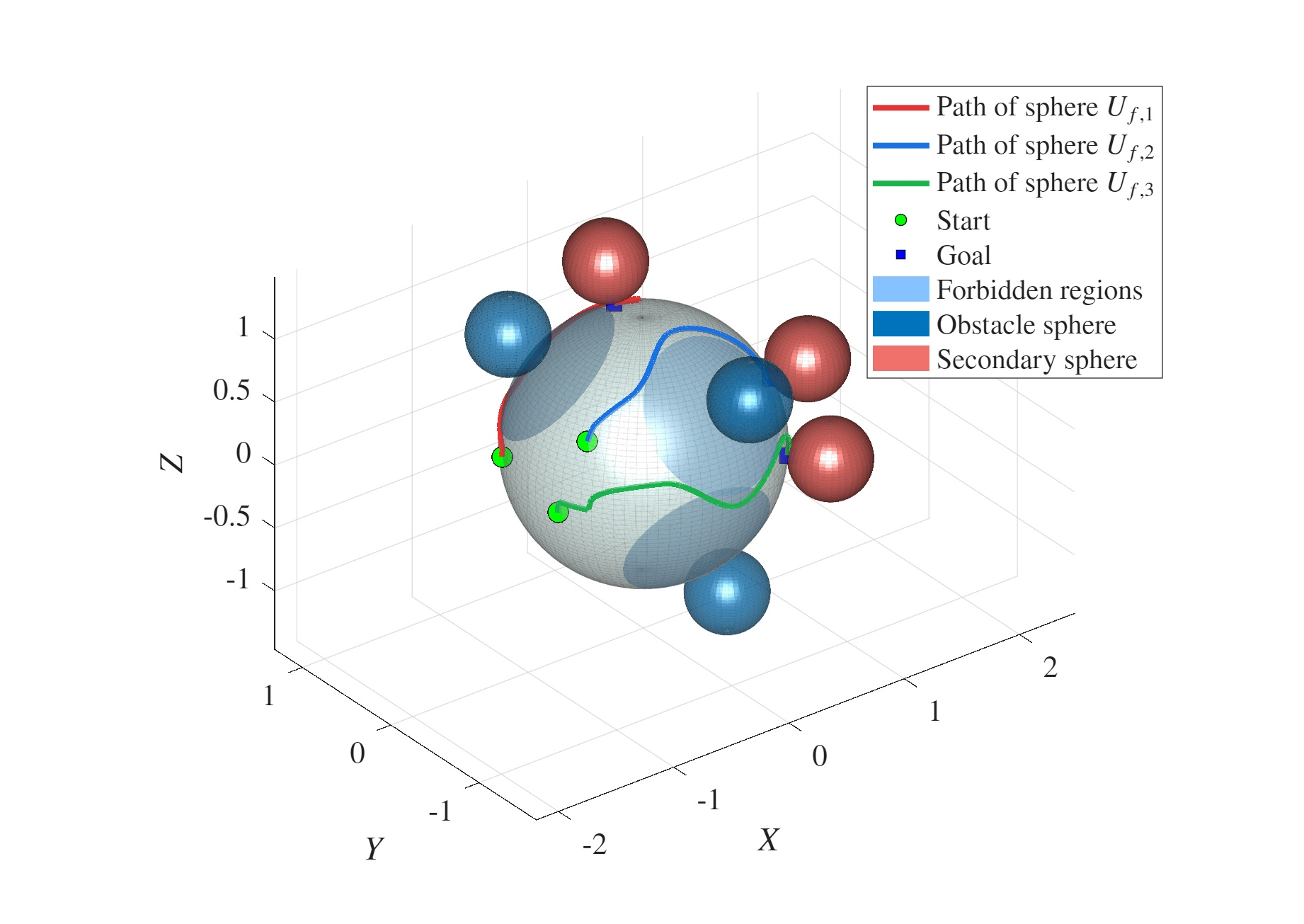}
        \caption{}\label{fig:res:b}
    \end{subfigure}
    \begin{subfigure}[t]{0.4\textwidth}
        \centering
        \includegraphics[width=\linewidth]{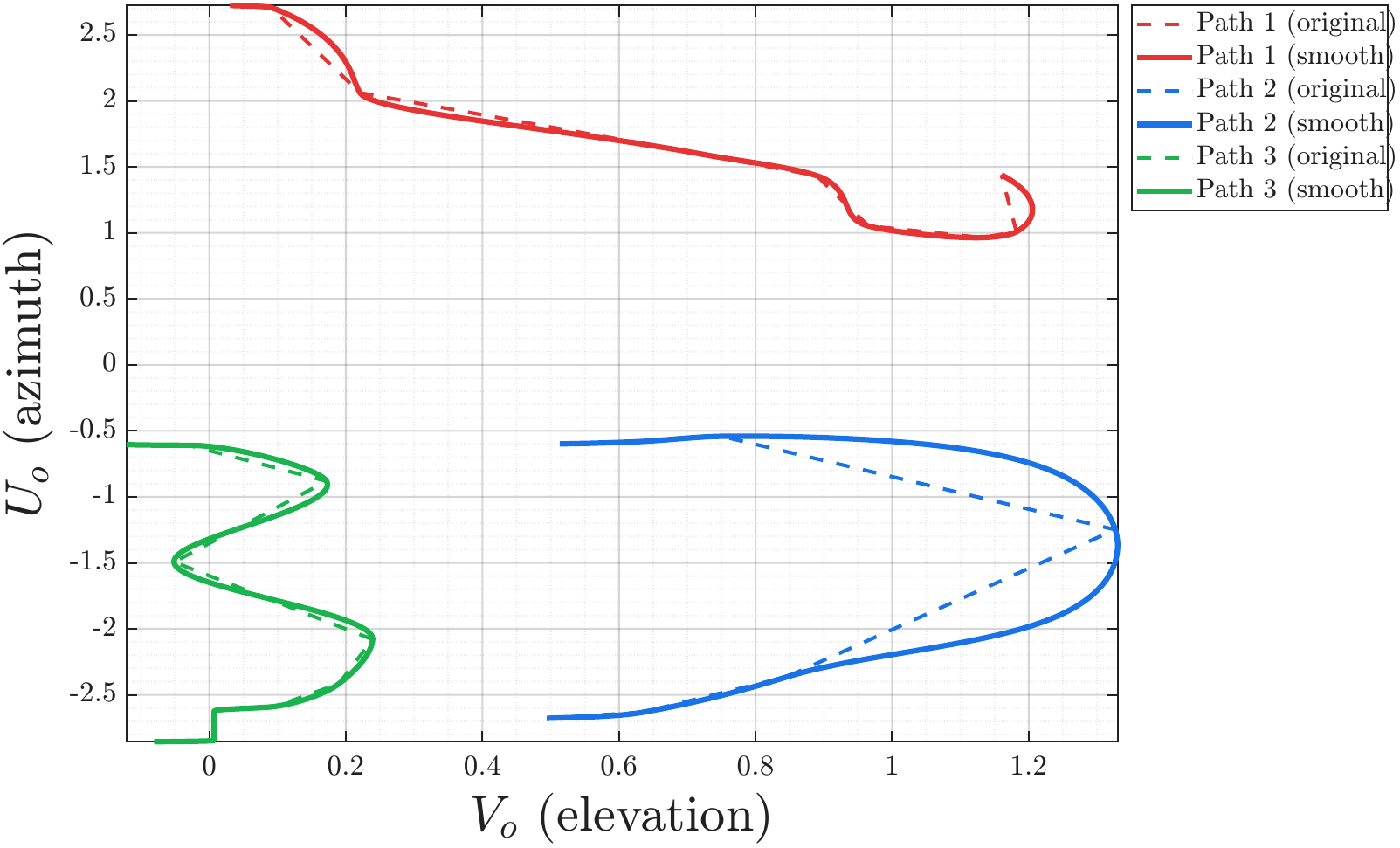}
        \caption{}\label{fig:res:c}
    \end{subfigure}
  \begin{subfigure}[t]{0.4\textwidth}
        \centering
        \includegraphics[width=\linewidth]{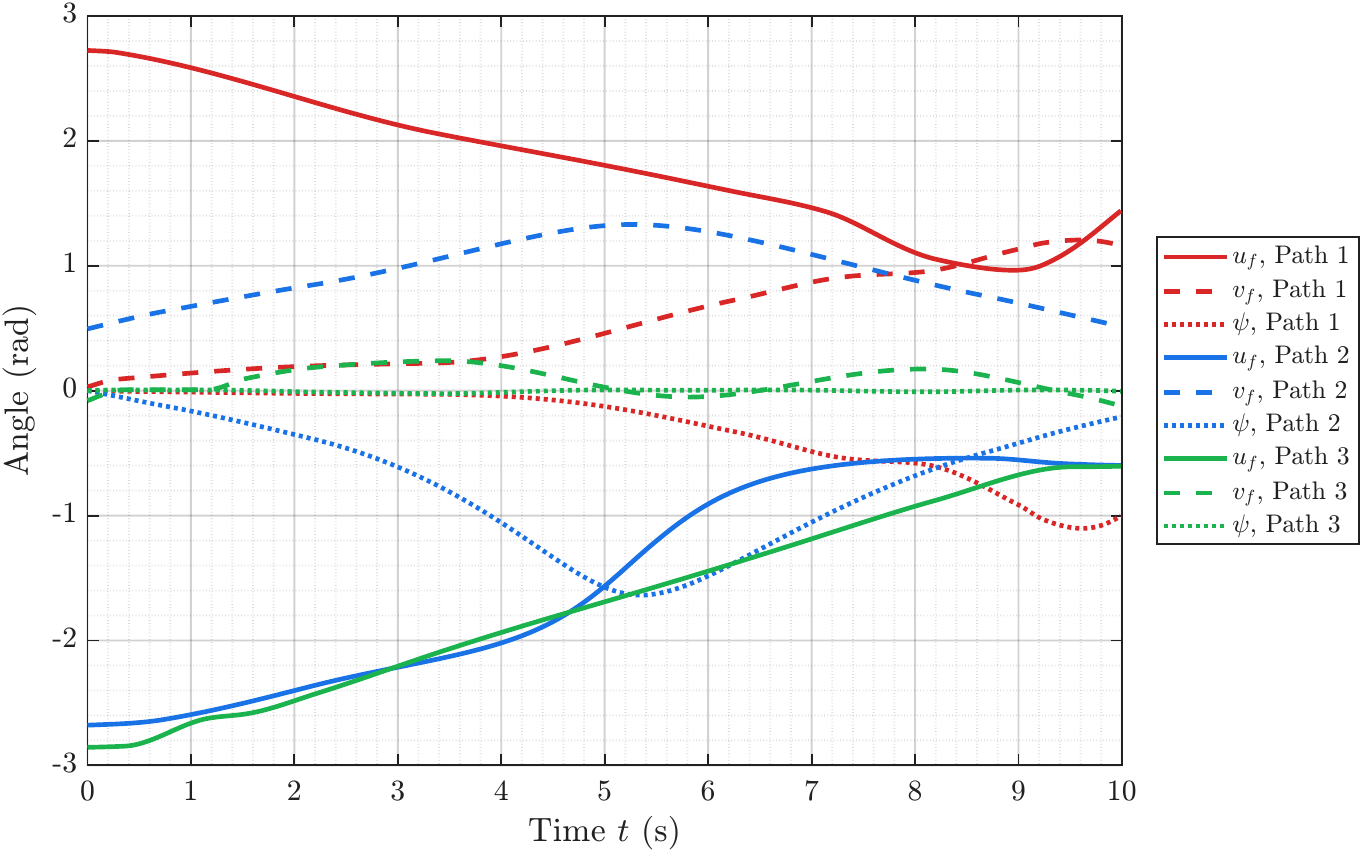}
        \caption{}\label{fig:res:c}
    \end{subfigure}
    \caption{Voronoi-based spherical-contact planning: (a) roadmap with inflated obstacle caps and paths for $U_{f,1}$--$U_{f,3}$; (b) 3D view with starts/goals; (c) $(v,u)$ projection comparing raw (dashed) and smoothed (solid) trajectories; (d) Time evolution of the reconstructed contact states $(u_f, v_f, \psi)$ for the three paths.}
    \label{fig:results:overview}
\end{figure*}

The remaining contact-state rates are then reconstructed linearly from the same rolling input:
\begin{equation}
    \begin{bmatrix}
        \dot{\bm{\eta}}_{f,i}\\
        \dot{\psi}_i
    \end{bmatrix}
    =
    \bm{g}_{r,i}(\bm{\chi}_i)\,\bm{\omega}_{xy,i}
    -
    \bm{e}_3\,\omega_{z,i},
    \label{eq:reconstruct_rest}
\end{equation}
where $\bm{g}_{r,i}(\bm{\chi}_i)\in\mathbb{R}^{3\times 2}$
collects the induced Montana coupling terms (secondary-side coordinate and spin-rate contributions) and
$\bm{e}_3=[0\; 0\; 1]^\top$ selects the $\omega_{z,i}$ component in the same convention used in
Section~\ref{sec:multi_sphere_rolling}. Together, Equation~\eqref{eq:omega_xy_inv} and~\eqref{eq:reconstruct_rest} provide
a closed reconstruction of $(\dot{\bm{\eta}}_{f,i}, \dot{\psi}_i)$ from the planned $\dot{\bm{\eta}}_{o,i}$.

To handle all contacts simultaneously in a unified reconstruction step, we stack $\dot{\bm{\eta}}_{o} := \mathrm{col}(\dot{\bm{\eta}}_{o,1}, \dots,\dot{\bm{\eta}}_{o,n})\in\mathbb{R}^{2n}$ and $\bm{\omega}_{xy} := \mathrm{col} (\bm{\omega}_{xy,1},\dots, \bm{\omega}_{xy,n})\in\mathbb{R}^{2n}$ to obtain
\begin{equation*}
    \bm{\omega}_{xy} = \bm{G}_o(\bm{\chi})^{-1}\,
    \dot{\bm{\eta}}_{o},
    \bm{G}_o(\bm{\chi}) := \mathrm{diag}\!\big(
    \bm{g}_{o,1}(\bm{\chi}_1),\dots,\bm{g}_{o,n}(\bm{\chi}_n)\big),
    \label{eq:stacked_inverse}
\end{equation*}
Applying \eqref{eq:reconstruct_rest} contact-wise and integrating the reconstructed rates yields the full stacked
contact trajectory $\bm{\chi}(t) = \mathrm{col}(\bm{\chi}_1(t),\dots,\bm{\chi}_n(t))$,
thereby satisfying the coupled no-slip constraints imposed by~\eqref{eq:multi_contact_main}. In implementation,
$\bm{g}_{o,i}(\bm{\chi}_i)$ is evaluated using the current integrated state $\bm{\chi}_i(t)$, with $\bm{\chi}_i(0)$
provided by the initial contact configuration.

\section{Results and Discussion}
\label{Sec:ResultsandDiscussion}
In this section, we demonstrate the proposed planner on a representative multi-contact scenario and evaluate its behaviour under obstacle constraints.

All simulations are implemented in MATLAB on an 11th Gen Intel(R) Core(TM) i5-11320H @ 3.20GHz PC with 16.0 GB RAM.The radius of the primary sphere is set to $R_o=1.0$, 
and three secondary spheres of radius $R_f=0.4$ remain 
in continuous contact with the host surface and roll 
without slip, following the coupled multi-contact kinematic 
model in~\eqref{eq:multi_contact_main}. The environment 
contains three spherical obstacles of physical radius 
$R_{\mathrm{obs}}=0.3$, whose induced forbidden regions 
on the unit sphere are modelled as inflated spherical caps 
as in~\eqref{eq:rho_eff_voronoi} and~\eqref{eq:seed_feasible_voronoi}. The hexagonal boundary radius follows from~\eqref{eq:delta_hex} as 
$\delta_{\mathrm{hex},k} \approx 0.827$\,rad. The spherical roadmap is then constructed from a
Voronoi tessellation seeded by best-candidate sampling~\eqref{eq:best_candidate_voronoi},
using $N_{\mathrm{seeds}}=200$ and candidate multiplier $M=30$. Collision checking along candidate edges is performed with angular discretization step $0.05$ rad, which sets the resolution for edge expansion and feasibility validation on the spherical manifold.

To validate the proposed framework, start and goal configurations are defined for the three secondary spheres in the presence of three spherical obstacles. The planning task is to compute collision-free multi-contact paths on the spherical surface. As shown in Fig.~\ref{fig:results:overview}, the Voronoi-based graph search successfully generates feasible paths connecting the boundary conditions. The resulting discrete shortest paths, shown in red, blue, and green, bypass the expanded obstacle regions highlighted in light blue and provide collision-free surface trajectories for all three contacts.

\begin{figure}[t!]
	\centering
	\includegraphics[width=3.4 in]{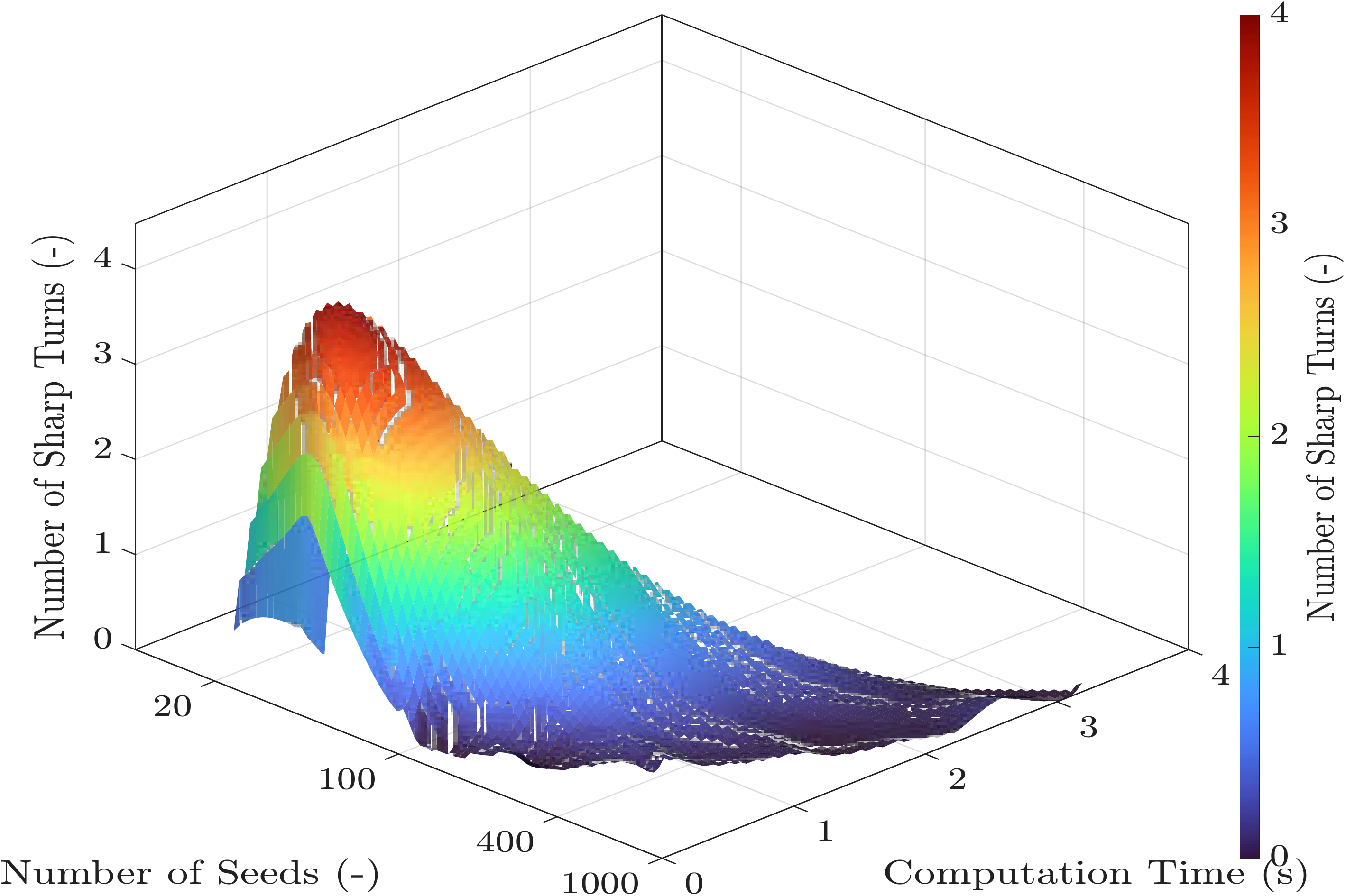}
	\caption{Trade-off surface between number of Voronoi seeds $N_{\mathrm{seeds}}$, computation time (s), and number of sharp turns in the planned path. Increasing seed density reduces path sharpness but incurs higher computation cost, with diminishing returns beyond $N_{\mathrm{seeds}} \approx 200$.}
 \label{Fig:surface——seeds}
\end{figure}

Fig.~\ref{fig:results:overview}(a) and Fig.~\ref{fig:results:overview}(b) show that the roadmap preserves connectivity over the admissible spherical surface even under relatively large inflated forbidden caps. The planned routes remain well separated from the obstacle boundaries, which is important in rolling systems because near-boundary motion reduces effective clearance and makes the reconstructed contact evolution more sensitive to local curvature and discretization. Fig.~\ref{fig:results:overview}(c) further shows that the raw graph paths are converted into smooth and realizable interpolation trajectories while preserving the overall route class around the obstacles. The change is particularly visible for Path 2 as 2nd secondary sphere, where the original discrete path contains a pronounced corner that is replaced by a more regular arc after interpolation. This is physically important because, for rolling bodies, abrupt changes of direction are difficult to realize under no-slip contact. Here, a sharp turn denotes a local directional change with angle approximately between $\pi/2$ and $\pi$ measured from consecutive trajectory directions; such turns are generally undesirable since they correspond to concentrated curvature changes that can lead to poor rolling realization and may push the reconstructed motion too close to forbidden regions.

This effect is quantified in Fig.~\ref{Fig:surface——seeds} with the purpose of generalisation by running simulations with over 40 case studies. Please note that this is study is done before doign any interpolation smoothing to evaluate the developed 2D manifold Voronoi search method. For low seed counts, the planner may return paths with up to about $4$ sharp turns while computation time remains below roughly $1$ s. Although such solutions are fast, they are too coarse for reliable rolling realization because the trajectory is dominated by graph corners rather than smooth manifold motion. As the seed number increases, the number of sharp turns drops rapidly. Around $N_{\mathrm{seeds}}\approx 200$, the planner reaches a practical regime where the number of sharp turns is close to zero or at most around one, while the runtime remains only about $1$--$1.5$ s. Beyond this point, increasing the seed count toward $400$--$1000$ raises the computation time to roughly $2$--$4$ s, but with only marginal improvement in path regularity. This indicates that $N_{\mathrm{seeds}}=200$ provides a reasonable trade-off between path quality and computational cost for the present framework.

The reconstructed rolling states in Fig.~\ref{fig:results:overview}(d) confirm that the interpolated surface trajectories are compatible with the inverse rolling reconstruction in \eqref{eq:omega_xy_inv} and \eqref{eq:reconstruct_rest}. Over the full $10$ s horizon, all reconstructed states remain continuous and bounded, with no visible discontinuities or oscillatory artifacts. Across the three contacts, the dominant coordinate trajectories vary approximately from $-2.8$ rad to $2.7$ rad, while the reconstructed spin remains within about $\pm 1.6$ rad. Path 2 exhibits the strongest transient around $t\approx4$--$5$ s, which is consistent with its larger geometric correction in Fig.~\ref{fig:results:overview}(c), yet the recovered states remain smooth and numerically well behaved. This shows that the geometric path quality directly affects the conditioning of the inverse rolling stage: better-cleared and more regular surface trajectories reduce the burden on reconstruction and improve the likelihood of physically realizable motion.

Overall, the results show that the proposed pipeline remains effective because it separates global obstacle avoidance from rolling-consistent motion reconstruction. The Voronoi roadmap resolves the topological routing directly on the spherical manifold, while the inverse reconstruction enforces kinematic consistency only after a smooth and realizable interpolation trajectory has been obtained. In reconfigurable rolling robots, this suggests a practical grouped strategy in which one contact group around a trapped object or host body is solved first, the forbidden and mutual-exclusion regions are updated, and the same procedure is then applied iteratively to the next group, consistent with the coordination perspective in \cite{tafrishi2025survey,zong2022kinematics}. On the manipulation side, the same planning structure is closely related to rolling-finger grasping, where fingertip contacts must move over the object surface while maintaining rolling compatibility. In that case, the main extension is the addition of secondary-side constraints such as fingertip workspace limits, preferred spin evolution, or task-specific terminal contact conditions. The present results therefore indicate that the proposed method can serve as a useful geometric and kinematic core for both modular rolling robots and rolling-contact grasping systems.

\label{sec: results}
\section{Conclusions}
\label{sec: conclusions}

This paper presented an obstacle avoidance framework for multi-contact path planning in spherical rolling robotics under no-slip constraints. We combined a clearance-favouring Voronoi roadmap constructed directly on the spherical 2D manifold with a reconstruction stage based on the generalized Montana multi-contact model. The planner accounts for embedded 3D obstacles (and mutual exclusion between rolling bodies) by representing forbidden regions as inflated spherical caps, reinforcing roadmap connectivity near obstacle boundaries via hex-ring insertion and local bridging, and producing manifold-consistent trajectories through intrinsic log--exp smoothing. The resulting smooth surface paths are then mapped to admissible multi-contact rolling motions by integrating the coupled Montana kinematics, avoiding the need to enforce the full nonholonomic coupling during the geometric search stage.

Simulation results with multiple secondary spheres and obstacle fields demonstrated that the proposed pipeline generates collision-free routes with robust clearance and yields coordinated rolling motions that remain consistent with the multi-contact kinematics. Across variations in contact number, obstacle layouts, and roadmap resolution, the method remained feasible and showed predictable trade-offs between sampling density, path quality, and computation. Future work will extend the framework beyond spherical host surfaces to general convex and non-convex geometries represented by meshes or point clouds, incorporate online replanning under dynamic obstacles and uncertainty in contact estimation, and validate the approach experimentally on spherical reconfigurable rolling robots and rolling-contact grasping devices.

\bibliographystyle{IEEEtran}
\bibliography{Reference} 
\end{document}